\documentclass[conference]{IEEEtran}
\IEEEoverridecommandlockouts
\usepackage{cite}
\usepackage{amsmath,amssymb,amsfonts}
\usepackage{algorithmic}
\usepackage{graphicx}
\usepackage{textcomp}
\usepackage{xcolor}
\usepackage{multirow}
\pdfoutput=1
\DeclareRobustCommand*{\IEEEauthorrefmark}[1]{%
  \raisebox{0pt}[0pt][0pt]{\textsuperscript{\footnotesize #1}}%
}

\def\BibTeX{{\rm B\kern-.05em{\sc i\kern-.025em b}\kern-.08em
    T\kern-.1667em\lower.7ex\hbox{E}\kern-.125emX}}
    
\begin{document}

\title{CFTrack: Enhancing Lightweight Visual Tracking through Contrastive Learning and Feature Matching}

\author{
    \IEEEauthorblockN{
        Juntao Liang\IEEEauthorrefmark{1}\textsuperscript{*}, 
        Jun Hou\IEEEauthorrefmark{2}\textsuperscript{*}, 
        Weijun Zhang\IEEEauthorrefmark{2}, 
        Yong Wang\IEEEauthorrefmark{1},
    }
    \IEEEauthorblockA{
        \IEEEauthorrefmark{1}School of Aeronautics and Astronautics, Shenzhen Campus of Sun Yat-sen University
        \IEEEauthorrefmark{2}Insta360 Research \\
        liangjt9@mail2.sysu.edu.cn,
        \{houjun, zhangweijun\}@insta360.com,
        \{wangyong5\}@mail.sysu.edu.cn
    }
    \thanks{\textsuperscript{*}These authors contribute equally to this work.}
}

\maketitle

\begin{abstract}

Achieving both efficiency and strong discriminative ability in lightweight visual tracking is a challenge, especially on mobile and edge devices with limited computational resources. Conventional lightweight trackers often struggle with robustness under occlusion and interference, while deep trackers, when compressed to meet resource constraints, suffer from performance degradation. To address these issues, we introduce CFTrack, a lightweight tracker that integrates contrastive learning and feature matching to enhance discriminative feature representations. CFTrack dynamically assesses target similarity during prediction through a novel contrastive feature matching module optimized with an adaptive contrastive loss, thereby improving tracking accuracy. Extensive experiments on LaSOT, OTB100, and UAV123 show that CFTrack surpasses many state-of-the-art lightweight trackers, operating at 136 frames per second on the NVIDIA Jetson NX platform. Results on the HOOT dataset further demonstrate CFTrack's strong discriminative ability under heavy occlusion. 

\end{abstract}


\section{Introduction}
\label{sec:intro}

Visual tracking, particularly single object tracking, is a fundamental task in computer vision. It has many applications in areas such as surveillance, human-computer interaction, and robotics. The process of visual tracking involves continuously identifying and locating a target within a video sequence. With the advances in deep learning, numerous trackers utilize Siamese networks for tracking \cite{10.1007/978-3-319-48881-3_56, 10.1007/978-3-030-58589-1_46}, which transform the tracking problem into learning a similarity score map between a template image and a candidate search region. These methods balance accuracy and processing speed, outperforming traditional approaches in many scenarios.

Despite these advances, challenges remain in deploying tracking systems on mobile or edge devices, such as drones and autonomous robots. These systems require real-time performance with minimal model parameters while exhibiting robust discriminative ability to handle challenges such as object disappearance, partial occlusion, and interference from similar objects. Trackers such as FEAR \cite{10.1007/978-3-031-20047-2_37} and MVT \cite{gopal2023mobile} tackle these issues by employing a dual-template strategy and integrating features using Mobile Vision Transformer, respectively. However, achieving high discriminative ability with compact models remains difficult, as smaller models often struggle to distinguish targets from similar backgrounds, particularly under occlusion or overlap. This limitation can result in inaccurate detection of target disappearance, which is critical for applications such as autonomous driving \cite{9259200} and drone navigation \cite{9561948}.

Robust feature representation is crucial to addressing challenges like occlusion, background clutter, and target appearance variation. Contrastive learning has proven effective for learning discriminative embeddings in many vision tasks, but its potential in lightweight tracking remains underexplored. Recent work, such as DRCI \cite{10219612}, demonstrates promising results in UAV tracking by using intra- and inter-video contrastive learning to improve feature discrimination during training. While DRCI achieves strong efficiency and precision, its contrastive learning module is only used during training and not during inference, reducing adaptability to dynamic environments. Many existing trackers focus on computational efficiency at the cost of representation quality, resulting in suboptimal performance in difficult scenarios.

In this work, we propose a novel framework that integrates contrastive learning and feature matching with a lightweight tracking pipeline (CFTrack) to achieve superior representation quality and discriminative ability. By incorporating a feature matching module and an adaptive contrastive loss function, CFTrack improves target-background separation and ensures temporal consistency in feature embeddings. Our approach explicitly optimizes the feature space to address distractors and appearance variations, delivering robust tracking while maintaining computational efficiency. Extensive experiments on benchmark datasets validate the effectiveness of CFTrack, showing significant gains in tracking performance. Our contributions are summarized as follows:

\begin{itemize}

\item We propose a contrastive feature matching module specifically designed for visual tracking, which leverages an adaptive contrastive learning to enhance target appearance discrimination. This module is integrated into the tracking framework to jointly optimize feature representations during tracking.

\item We develop a lightweight tracker, CFTrack, that balances high accuracy with computational efficiency by integrating the proposed contrastive feature matching module.

\item We conduct extensive evaluations on LaSOT, OTB100, UAV123 and HOOT datasets. Our results demonstrate that CFTrack outperforms many state-of-the-art lightweight trackers in accuracy and robustness, maintaining real-time inference speeds exceeding 136 $fps$ on the NVIDIA Jetson NX platform.

\end{itemize}

\section{Related Works}

\subsection{Lightweight Visual Tracking}

Lightweight visual tracking has advanced significantly with the rise of Siamese-based methods. Early methods, such as SiamFC \cite{10.1007/978-3-319-48881-3_56} and ECO \cite{Danelljan_2017_ECO}, achieve real-time performance on edge devices. However, their tracking accuracy remains inferior compared to state-of-the-art trackers. Recently, more efficient trackers have been introduced. Yan \emph{et al.} \cite{Yan_2021_CVPR} employ neural architecture search (NAS) to design efficient and compact models for resource-constrained hardware. Borsuk \emph{et al.} \cite{10.1007/978-3-031-20047-2_37} introduce a dual-template-based lightweight tracker, combining static and dynamic templates to handle appearance changes. Gopal \emph{et al.} \cite{gopal2023mobile} use Mobile Vision Transformer to integrate features from both the template and search regions, improving localization and classification for tracking. These advancements underline the focus on creating robust, real-time tracking solutions suitable for practical applications on mobile platforms. However, these methods often compromise on discriminative ability due to their simplified architectures.

\subsection{Contrastive Learning}
Contrastive learning is a self-supervised technique that aims to distinguish between positive pairs (augmentations of the same instance) and negative pairs (augmentations of different instances). It has shown promise in various computer vision tasks by learning effective feature representations \cite{chen2020simple, he2020momentum, Zhang_2021_CVPR}. In the filed of visual tracking, contrastive learning has also been employed to improve instance and category-level representation. Pi \emph{et al.} \cite{10.1007/978-3-031-20047-2_25} leverage instance-aware and category-aware modules combined with video-level contrastive learning to enhance inter-instance separability and intra-instance compactness. Wu \emph{et al.} \cite{Wu_2021_CVPR} employ contrastive learning in an unsupervised framework, using noise-robust temporal mining strategies to generate positive and negative pairs, thereby improving temporal consistency. While these methods improve robustness, they lack a focus on lightweight design, which is crucial for real-time applications. Dan \emph{et al.} \cite{10219612} specifically targets UAV tracking by utilizing contrastive learning within video frames to enhance feature discrimination against occlusion. However, the contrastive module is not used during inference, limiting its ability to distinguish targets from distractors in real-world tracking scenarios. In this paper, we address these gaps by incorporating a contrastive learning-based feature matching module that operates during both training and inference within a lightweight Siamese framework, ensuring accurate and efficient tracking even in resource-constrained applications.

\section{Methods}

\begin{figure*}[!htbp]
    \centering 
    \includegraphics[width=0.9\textwidth]{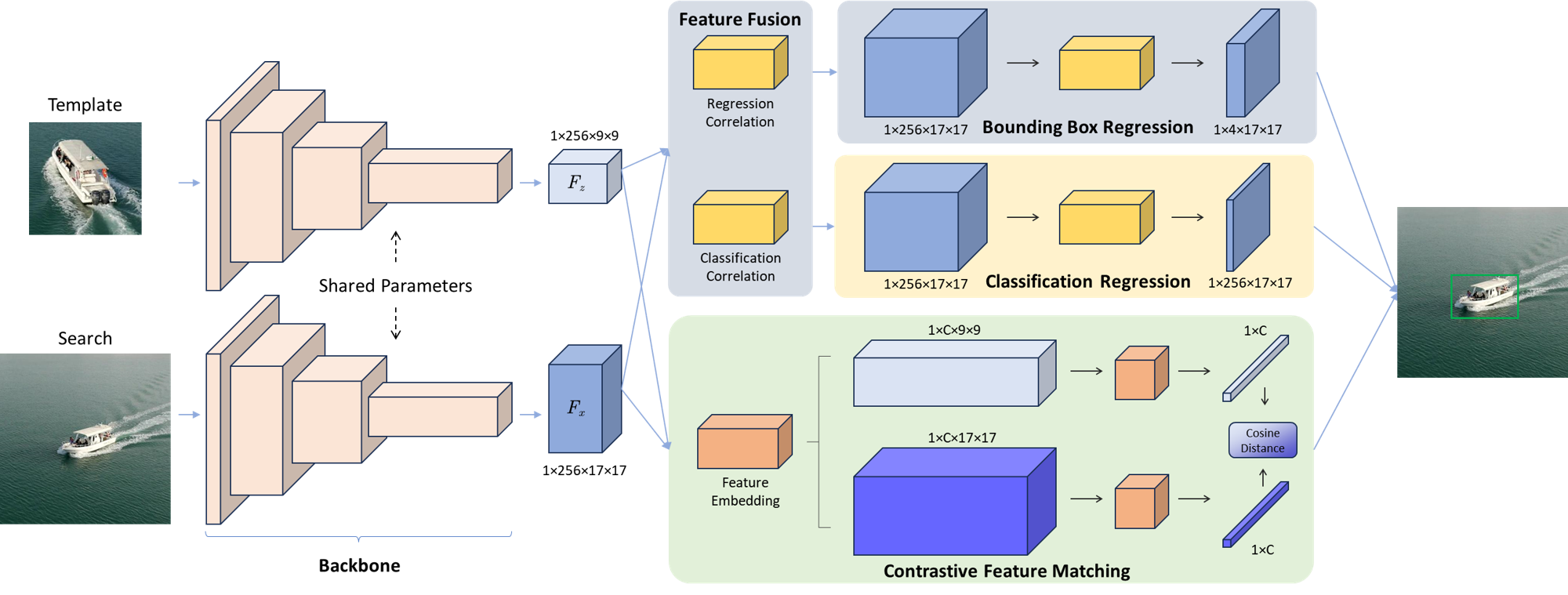}
    \caption{Architecture of the proposed CFTrack framework. The CFTrack framework consists of three key components: (1) a lightweight backbone for feature extraction, (2) correlation blocks for feature fusion, and (3) a prediction head including three branches for bounding box regression, classification, and contrastive feature matching.}
    \label{fig:fig1}
\end{figure*}

In this section, we present the pipeline of the proposed algorithm, which employs a Siamese structure for object tracking, as depicted in Fig. \ref{fig:fig1}. 

\subsection{Feature Extraction and Fusion}

We use MobileNetV2\cite{Sandler_2018_CVPR} for feature extraction with a Siamese network architecture. This network has two branches that share parameters: the template branch and the search branch. These branches take patches from the template (\( z \)) and search region (\( x \)), respectively. Once the backbone extracts features from the template and search region, a fusion module integrates these features. The fusion module, inspired by Alpha-Refine \cite{Yan_2021_CVPR_Alpha_Refine}, combines pixel-wise correlation with channel attention, effectively merging the features from both branches.

After feature extraction, the template and search feature maps, denoted as \( F_z \) and \( F_x \), are flattened along the channel dimension into vectors \( \mathbf{v}_z \) and \( \mathbf{v}_x \). Fusion is achieved by computing the element-wise product of these vectors:

\begin{equation}
\mathbf{v}_{\text{fused}} = \mathbf{v}_z \odot \mathbf{v}_x.
\end{equation}

This fused vector is reshaped into a pixel-wise fused feature map, followed by channel-wise attention. Average pooling is applied to each channel to produce a vector \( \mathbf{v} \), which is processed through two fully connected layers with activation functions:

\begin{equation}
\mathbf{v}' = \sigma(\mathbf{W}_2 \cdot \text{ReLU}(\mathbf{W}_1 \cdot \mathbf{v} + \mathbf{b}_1) + \mathbf{b}_2),
\end{equation}
where \( \sigma \) is the activation function, and \( \mathbf{W}_1 \), \( \mathbf{W}_2 \), \( \mathbf{b}_1 \), and \( \mathbf{b}_2 \) are the weights and biases of the fully connected layers. The output vector \( \mathbf{v}' \) is then multiplied with the feature map to apply channel-wise attention. The final fused feature map retains the spatial dimensions of the search region's feature map and contains 81 channels.

\subsection{Contrastive Feature Matching Module}
The Contrastive Feature Matching (CFM) module is specifically designed to assess target occlusion and disappearance by evaluating the feature similarity, enhancing discriminative ability in lightweight trackers. Unlike conventional feature matching approaches, the CFM module integrates contrastive learning principles with an adaptive margin mechanism to dynamically address similar distractors (hard negatives) within the tracking framework.

The CFM module processes the feature maps to generate embeddings that encapsulate the key characteristics of the target. The template and search area features are input into an embedding module comprising convolutional and activation layers with shared parameters. This module outputs 256-dimensional feature vectors, $\mathbf{v}_z^{\text{embed}}$ and $\mathbf{v}_x^{\text{embed}}$, representing the target's appearance in the template and the most confident location in the search area, respectively.

We compute the cosine similarity between these vectors to assess the appearance similarity:

\begin{equation} \text{Similarity} = \frac{\mathbf{v}_z^{\text{embed}} \cdot \mathbf{v}_x^{\text{embed}}}{|\mathbf{v}_z^{\text{embed}}| |\mathbf{v}_x^{\text{embed}}|}. \end{equation}

This similarity measure is then combined with the classification score to produce a final confidence score for tracking the target:

\begin{equation} \text{Confidence} = \max(\text{Similarity} \cdot \text{Score}, 0), \end{equation}
which reflects the tracker's discriminative ability and serves as an indicator of target presence.

\subsubsection{Targeted Data Sampling Strategy}

To support the contrastive learning process, we implement a targeted data sampling strategy that ensures the effective generation of positive and negative pairs:

\begin{itemize}
    \item \textbf{Positive Pair:} Samples of the same target taken from different frames of a video sequence, labeled as \(y=1\) (similar).
    \item \textbf{Negative Pair:} Samples taken from different video sequences, labeled as \(y=0\) (dissimilar).
\end{itemize}

For positive pairs (\(y=1\)), the objective is to minimize the cosine distance \(D\), bringing their feature representations closer together. For negative pairs (\(y=0\)), the goal is to maximize the distance, pushing their representations apart.

\subsubsection{Adaptive Margin Mechanism}

The CFM module incorporates an adaptive margin mechanism to dynamically adjust the separation between positive and negative pairs during training. The margin is determined based on the cosine similarity, with the cosine distance \(D\) defined as:

\begin{equation}
D = 1 - \text{Similarity}.
\end{equation}

The adaptive margin \(m(D)\) is then expressed as:

\begin{equation} m(D) = m_0 + \beta \cdot e^{-\gamma D}, \end{equation}
where \(m_0\) is the base margin, and \(\beta,\gamma\) being hyperparameters controlling the margin adaptation rate. This function increases the margin for negative pairs with higher similarity, thus penalizing them more during training.

\subsubsection{Adaptive Contrastive Loss Function}
The adaptive margin mechanism is integrated into the adaptive contrastive loss function \(L_{adapt}\), which optimizes the embeddings produced by the CFM module:
\begin{equation}
\begin{split}
L_{adapt} = y \cdot D^2 + (1 - y) \cdot \left[\max(0, m(D) - D)\right]^2.
\end{split}
\end{equation}

The CFM module is trained jointly with other components of the tracker, adopting a multi-task learning strategy. In this setup, contrastive learning serves as an auxiliary task, regularizing the training process and reinforcing the primary tracking objective.

\subsection{Prediction Head}

The prediction head includes three branches: classification, bounding box regression, and the proposed CFM module. The classification branch predicts whether the target is present, while the bounding box regression branch estimates the target's precise size and location.

Both the classification and bounding box regression branches receive the fused feature maps. These branches consist of multiple \( 3 \times 3 \) and \( 5 \times 5 \) depth-wise separable convolutions, which output classification scores and bounding box coordinates for each anchor point.

\subsection{Loss Function}

During the training, we employ different loss functions to optimize the outputs of both the classification and bounding box regression branches of the model. For classification, we apply a weighted Binary Cross-Entropy (BCE) loss, denoted as \(L_{\text{cls}}\). For bounding box regression, we apply the smooth L1 loss, denoted as \(L_1\), which is less sensitive to outliers compared to the standard L1 loss.

Finally, the loss function is defined as:

\begin{equation}
L_{total}= \lambda_1L_{cls} + \lambda_2 L_1 + \lambda_3 L_{adapt},
\end{equation}
where $\lambda_1$ , $\lambda_2$ and $\lambda_3$ are set to 1.0, 1.0, and 2.0.

\section{Experimental Evaluation}

In this section, we present the implementation details of our proposed tracker and compare its performance with other related lightweight trackers. Additionally, we discuss the results of the ablation study conducted on the proposed contrastive feature matching module.

\subsection{Implementation Details}

The proposed method is implemented using Python 3.8 and PyTorch 1.10.1. All experiments are conducted on the system equipped with an Intel i9 CPU, an NVIDIA 3090 GPU, and running Ubuntu 18.04.

The proposed tracker is trained on the training splits of LaSOT \cite{Fan_2019_CVPR}, COCO \cite{10.1007/978-3-319-10602-1_48}, GOT-10k \cite{8922619}, and TrackingNet \cite{Muller_2018_ECCV} datasets. Data augmentations, including horizontal flips and brightness jittering, are applied to increase data diversity. The input template image is resized to 144×144 pixels, while the search image is resized to 272×272 pixels. The network is optimized using the AdamW optimizer with a weight decay of 1e-4. Training is conducted over 300 epochs, with 60,000 samples processed per epoch. The initial learning rate is 1e-4, decreasing by a factor of 10 every 40 epochs after 160 epochs. We set the hyperparameters \(m_0, \beta\) and \(\gamma\) in Eq 6. to 1.0, 1.0, and 2.0, respectively.

\begin{table}[!ht]
\centering
\caption{State-of-the-art comparison on three benchmarks in terms of AUC score (in \%). The deep trackers are in the upper part of the table, while the lightweight trackers are in the lower part. The best score is highlighted in {\color[HTML]{0000FF} blue} while the best lightweight score is highlighted in {\color[HTML]{FF0000} red}.}
\label{tab: tabel1}
\resizebox{\columnwidth}{!}{
\begin{tabular}{ccccc}
\hline
Trackers       & Source   & OTB100                        & UAV123                       & LaSOT   \\ \hline
ATOM           & CVPR'19  & 66.3                         & 63.2                         & 51.4      \\
Ocean(offline) & ECCV'20  & 67.6                         & 57.4                         & 50.5      \\
STARK          & ICCV'21  & 68.1                         & 68.2                         & 67.1      \\
MixFormer      & CVPR'22  & {\color[HTML]{0000FF} 69.6}  & 68.7                         & 67.9      \\
SeqTrack-B256  & CVPR'23  & 69.1                        & {\color[HTML]{0000FF} 69.2}   & {\color[HTML]{0000FF} 69.9}  \\ \hline
ECO            & CVPR'17  & 69.1                         & 53.2                         & 32.4                         \\
LightTrack     & CVPR'21  & 66.2                         & 62.5                         & 52.4                         \\
HiFT           & ICCV'21  & 58.9                         & 58.9                         & -                            \\
FEAR-XS        & ECCV'22  & 66.1                         & -                            & 53.5                         \\
TCTrack        & CVPR'22  & 62.4                         & 60.4                         & -                            \\
TCTrack++      & PAMI'23  & 66.2                         & 60.8                         & -                            \\
HiT-Tiny       & ICCV'23  & 54.3                         & 58.7                         & 54.8                         \\
MVT            & BMCV'23  & 67.3                         & 59.8                         & 55.3                         \\
E.T.Track      & WACV'23  & 68.1                         & 62.7                         & {\color[HTML]{FF0000} 59.7}  \\
\textbf{CFTrack} & \textbf{Ours} & {\color[HTML]{FF0000} 69.3} & {\color[HTML]{FF0000} 64.4} & 56.0                    \\ \hline
\end{tabular}}
\end{table}

\subsection{Comparison with State-of-the-art Trackers}

We compare the CFTrack with 14 state-of-the-art trackers, including both lightweight trackers \cite{Danelljan_2017_ECO, Yan_2021_CVPR, Cao_2021_ICCV_HiFT, 10.1007/978-3-031-20047-2_37, Cao_2022_CVPR, 10225683, Kang_2023_ICCV, gopal2023mobile, Blatter_2023_WACV} and deep trackers \cite{Danelljan_2019_CVPR_ATOM, 10.1007/978-3-030-58589-1_46, Yan_2021_ICCV, Cui_2022_CVPR, Chen_2023_CVPR}, across three tracking benchmarks, including LaSOT \cite{Fan_2019_CVPR}, OTB100 \cite{7001050} and UAV123 \cite{10.1007/978-3-319-46448-0_27}. Tracking performance is evaluated with the area-under-curve (AUC) metric, which measures the area under the success rate curve, representing the average proportion of successful frames across varying Intersection over Union (IoU) thresholds from 0 to 1. Detailed results are presented in Table \ref{tab: tabel1}. 

\textbf{OTB100.} OTB100 includes 100 sequences with diverse tracking challenges, serving as a widely recognized benchmark for assessing general tracking performance. On OTB100, CFTrack attains an AUC of 69.3\%, which is almost equivalent to that of MixFormer. Moreover, CFTrack runs 4.6 times faster than MixFormer, underscoring its computational efficiency and suitability for real-time applications.

\textbf{UAV123.} UAV123 is a large-scale aerial tracking benchmark comprising 123 sequences with over 112K frames. Performance evaluation on UAV123 verifies the tracker's effectiveness in common aerial tracking conditions. CFTrack attains an AUC of 64.4\% on UAV123, outperforming UAV-specific trackers such as TCTrack and TCTrack++. This result demonstrates CFTrack's adaptability in domain-specific conditions common to aerial tracking scenarios.

\textbf{LaSOT.} LaSOT is a large-scale benchmark designed for long-term tracking, containing 1,400 sequences. On LaSOT, CFTrack attains an AUC of 56.0\%, outperforming multiple lightweight methods such as MVT and HiT-Tiny, while ranking just behind E.T.Track. Notably, CFTrack runs 3.6 times faster than E.T.Track. Although certain deep trackers offer superior performance, their gains typically come at the cost of increased computational complexity and reduced processing speed.

\textbf{Computational Efficiency.} We conduct a comparative analysis of the parameters and speed metrics of our tracker against other state-of-the-art trackers. As shown in Table \ref{tab: tabel2}, CFTrack offers a favorable trade-off between model complexity and speed. With 4.167 million parameters and 1.24 GFLOPs, CFTrack operates at 368 $fps$, placing it among the fastest lightweight models while maintaining competitive accuracy. This high efficiency can be attributed to its hardware-friendly design and operator-level optimizations, which facilitate effective parallelization and reduce memory overhead.

\begin{table}[!ht]
\centering
\caption{Comparison of FLOPs, model size (parameters) and speed on NVIDIA 3090 GPU with related lightweight trackers, speed in running frames-per-second ($fps$). The deep trackers are in the upper part of the table, while the lightweight trackers are in the lower part.}
\label{tab: tabel2}
\resizebox{\columnwidth}{!}{
\begin{tabular}{ccccc}
\hline
Trackers            & Source        & Parameters & FLOPs  & Speed \\ \hline
STARK               & ICCV'21       & 47.2M      & 20.4G  & 88   \\
MixFormer           & CVPR'22       & 35.61M     & 23.04G & 79   \\
SeqTrack-B256       & CVPR'23       & 85.65M     & 65.74G & 73  \\ \hline
LightTrack          & CVPR'21       & 1.97M      & 0.54G  & 138 \\
FEAR-XS             & ECCV'22       & 1.37M      & 0.48G  & 275 \\
TCTrack             & CVPR'22       & 8.5M       & 6.9G   & 225 \\
TCTrack++           & PAMI'23       & 7.08M      & 4.38G  & 199 \\
HiT-Tiny            & ICCV'23       & 9.60M      & 0.98G  & 386 \\
MVT                 & BMCV'23       & 5.49M      & 0.77G  & 236 \\
E.T.Track           & WACV'23       & 6.98M      & 1.56G  & 101 \\
\textbf{CFTrack}    & \textbf{Ours} & 4.33M      & 1.44G  & 368 \\ \hline
\end{tabular}}
\end{table}

\subsection{Qualitative Evaluation}

\begin{figure}[!htbp]
    \centering 
    \includegraphics[width=0.5\textwidth]{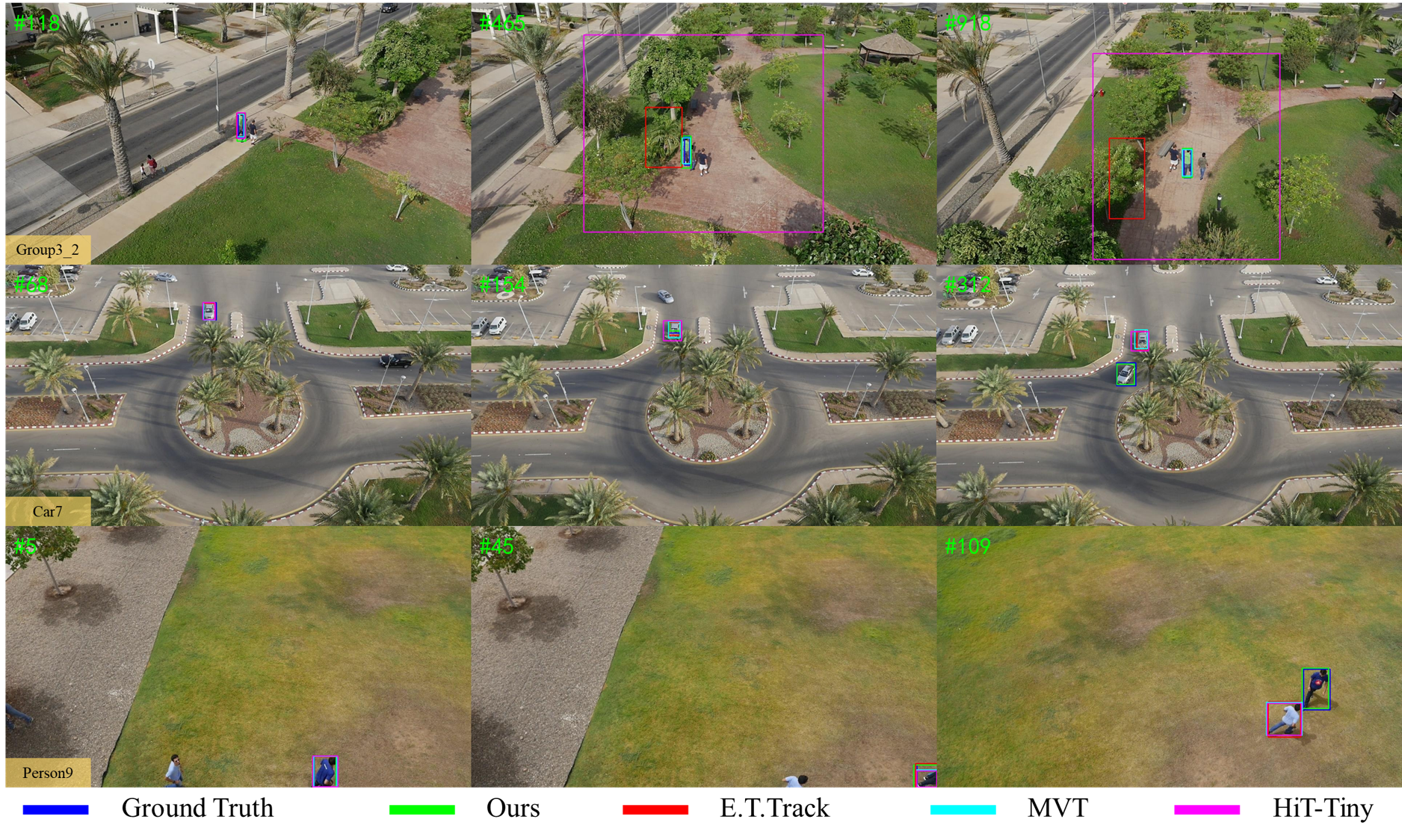}
    \caption{Qualitative comparison results of our CFTrack with other three lightweight trackers on UAV123 (Zoom in for better view).}
    \label{fig:fig4}
\end{figure}

We present qualitative tracking results for CFTrack in challenging occlusion scenarios using the UAV123 dataset, as illustrated in Fig. \ref{fig:fig4}. These results emphasize the robustness of our method compared to other state-of-the-art lightweight trackers, particularly under conditions of heavy occlusion, target disappearance, and environmental clutter. CFTrack demonstrates superior performance, ensuring discriminative and consistent tracking. A video of the qualitative experiment is provided in the supplementary material.

\subsection{Ablation Study}

\subsubsection{Effect of Contrastive Feature Matching}

\begin{table}[!ht]
\centering
\caption{Comparison of AUC score (in \%), FLOPs and model size (parameters) between the proposed CFTrack and the baseline method on three benchmarks.}
\label{tab: tabel4}
\resizebox{\columnwidth}{!}{
\begin{tabular}{c|ccccc}
\hline
Method       & OTB100 & UAV123 & LaSOT & Parameters & FLOPs  \\ \hline
Baseline     & 65.5   & 59.5   & 51.2  & 3.90M    & 1.31G \\
Baseline+CFM & 69.3   & 64.4   & 56.0  & 4.33M    & 1.44G \\ \hline
\end{tabular}}
\end{table}

To evaluate the effectiveness of the proposed CFM module in enhancing tracking performance, we conduct ablation experiments on three benchmarks, using the structure in Fig. \ref{fig:fig1} without the CFM module as a baseline. The comparisons are shown in Table \ref{tab: tabel4}.  We can see that the proposed CFM module improves the AUC by 4.2\% on average across all datasets, with notable gains of 4.8\% on LaSOT, 3.8\% on OTB100, and 4.9\% on UAV123. By leveraging contrastive learning, the CFM module enhances feature representation by optimizing the feature space for improved target-background separation and temporal consistency. Furthermore, the adaptive loss function refines feature representation by focusing on hard negative samples, dynamically adjusting margins to separate negatives while clustering positives, resulting in improved tracking success of the tracker.

\subsubsection{Discriminative Ability}

To investigate the effect of proposed CFM module in enhancing the tracker's discriminative ability, we perform ablation experiments using the HOOT \cite{sahin2023hoot} dataset, which is designed for scenarios with heavy occlusion (median 68\% occlusion per video). An online evaluation method is used, where the tracker is reinitialized upon losing the target (IoU below 80\%), and confidence scores are analyzed under varying visibility conditions.

\begin{figure}[!htbp]
    \centering 
    \includegraphics[width=0.5\textwidth]{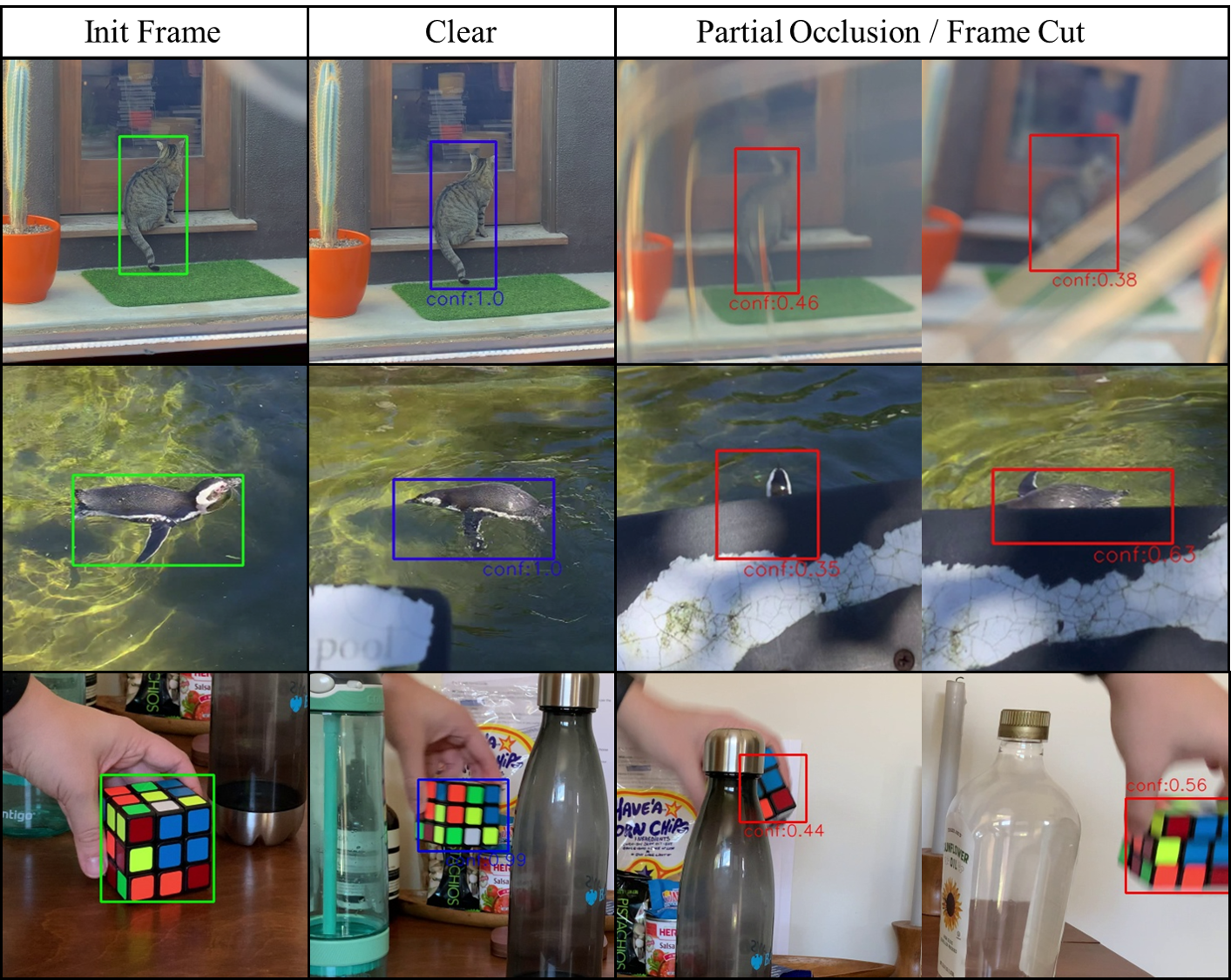}
    \caption{Qualitative results under different visibility conditions. Red/blue bounding boxes indicate confidence below/above 0.8.}
    \label{fig:fig2}
\end{figure}

\begin{table}[!ht]
\centering
\caption{Ablation study of the impact of CFM for tracker's discriminative performance (average confidence on HOOT test set) under different visibility conditions. "Occ." for occlusion and "Frame Cut" for target partially exiting the frame.}
\label{tab: tabel3}
\resizebox{\columnwidth}{!}{
\begin{tabular}{c|ccccc}
\hline
Method & Full Occ. & Partial Occ. & Frame Cut & Absent & Clear\\ \hline
Baseline    & 0.62 & 0.92 & 0.95 & 0.93 & 0.96 \\
Baseline+CFM & 0.47 & 0.72 & 0.79 & 0.65 & 0.88 \\ \hline
\end{tabular}}
\end{table}

As shown in Table \ref{tab: tabel3} and Fig. \ref{fig:fig2}, the CFM module improves discrimination under challenging conditions. Confidence decreases more sharply with CFM in full occlusion (0.47 vs. 0.62), partial occlusion (0.72 vs. 0.92), and target absence (0.79 vs. 0.93), indicating better target-background distinction. Under clear visibility, confidence slightly decreases (0.88 vs. 0.96), reflecting heightened sensitivity. These results confirm that the CFM module enhances the tracker's robustness by reducing false positives in occlusion scenarios.

\subsection{Real-World Tests}

\begin{figure}[!htbp]
    \centering
    \includegraphics[width=0.5\textwidth]{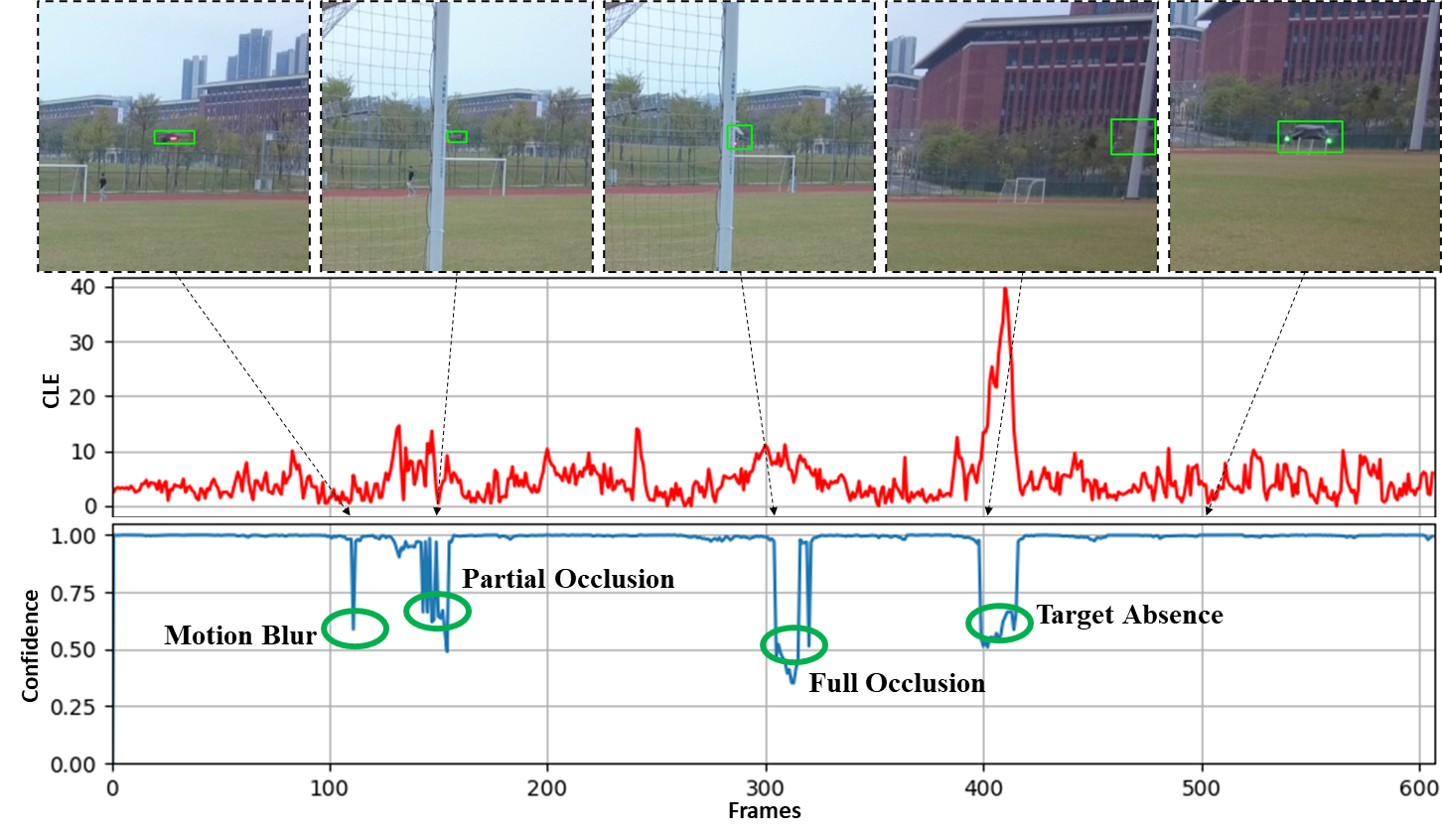}
    \caption{Real-world test results. The tracking results are marked with \textcolor{green}{green} and the CLE represents the center location error.}
    \label{fig:fig3}
\end{figure}

To assess the practical applicability of our method, we deploy CFTrack on the NVIDIA Jetson NX 8GB platform using TensorRT, optimized for mobile and edge applications. We collect the videos of drones and perform tracking on the platform. Our tests cover challenging scenes including occlusion, target absence, and motion blur. Tracking accuracy is evaluated using the center location error (CLE), with a 20-pixel threshold. As shown in Fig. \ref{fig:fig3}, CFTrack demonstrates strong accuracy and discriminative ability when facing occlusion. Additionally, it achieves an average speed of 136 $fps$, highlighting its suitability for real-time applications on edge devices.

\section{Conclusion}

In this work, we introduce CFTrack to address discriminative challenges in lightweight visual tracking. Our method enhances the quality of learned feature representations by integrating a contrastive feature matching module, thereby improving the discriminative ability in tracking. Our approach achieves better performance on benchmarks including LaSOT, OTB100, UAV123, and HOOT. CFTrack is among the first to combine these techniques, enhancing both accuracy and efficiency in real-time, resource-constrained environments. We believe this plug-and-play feature matching module can benefit many lightweight trackers and inspire further innovations in the field.

\bibliographystyle{IEEEbib}
\bibliography{refs}

\begin{thebibliography}{10}

\bibitem{10.1007/978-3-319-48881-3_56}
Luca Bertinetto, Jack Valmadre, Jo{\~a}o~F. Henriques, Andrea Vedaldi, and Philip H.~S. Torr,
\newblock ``Fully-convolutional siamese networks for object tracking,''
\newblock in {\em ECCV 2016 Workshops}, 2016, pp. 850--865.

\bibitem{10.1007/978-3-030-58589-1_46}
Zhipeng Zhang, Houwen Peng, Jianlong Fu, Bing Li, and Weiming Hu,
\newblock ``Ocean: Object-aware anchor-free tracking,''
\newblock in {\em ECCV}, 2020, pp. 771--787.

\bibitem{10.1007/978-3-031-20047-2_37}
Vasyl Borsuk, Roman Vei, Orest Kupyn, Tetiana Martyniuk, Igor Krashenyi, and Ji{\v{r}}i Matas,
\newblock ``Fear: Fast, efficient, accurate and robust visual tracker,''
\newblock in {\em ECCV}, 2022, pp. 644--663.

\bibitem{gopal2023mobile}
Goutam~Yelluru Gopal and Maria Amer,
\newblock ``Mobile vision transformer-based visual object tracking,''
\newblock in {\em BMVC}, 2023.

\bibitem{9259200}
João~Eduardo Hoffmann, Hilkija~Gaïus Tosso, Max Mauro~Dias Santos, João~Francisco Justo, Asad~Waqar Malik, and Anis~Ur Rahman,
\newblock ``Real-time adaptive object detection and tracking for autonomous vehicles,''
\newblock {\em IEEE Transactions on Intelligent Vehicles}, vol. 6, no. 3, pp. 450--459, 2021.

\bibitem{9561948}
Zhichao Han, Ruibin Zhang, Neng Pan, Chao Xu, and Fei Gao,
\newblock ``Fast-tracker: A robust aerial system for tracking agile target in cluttered environments,''
\newblock in {\em ICRA}, 2021, pp. 328--334.

\bibitem{10219612}
Dan Zeng, Mingliang Zou, Xucheng Wang, and Shuiwang Li,
\newblock ``Towards discriminative representations with contrastive instances for real-time uav tracking,''
\newblock in {\em ICME}, 2023, pp. 1349--1354.

\bibitem{Danelljan_2017_ECO}
Martin Danelljan, Goutam Bhat, Fahad Shahbaz~Khan, and Michael Felsberg,
\newblock ``Eco: Efficient convolution operators for tracking,''
\newblock in {\em CVPR}, 2017.

\bibitem{Yan_2021_CVPR}
Bin Yan, Houwen Peng, Kan Wu, Dong Wang, Jianlong Fu, and Huchuan Lu,
\newblock ``Lighttrack: Finding lightweight neural networks for object tracking via one-shot architecture search,''
\newblock in {\em CVPR}, 2021, pp. 15180--15189.

\bibitem{chen2020simple}
Ting Chen, Simon Kornblith, Mohammad Norouzi, and Geoffrey Hinton,
\newblock ``A simple framework for contrastive learning of visual representations,''
\newblock in {\em International conference on machine learning}. PMLR, 2020, pp. 1597--1607.

\bibitem{he2020momentum}
Kaiming He, Haoqi Fan, Yuxin Wu, Saining Xie, and Ross Girshick,
\newblock ``Momentum contrast for unsupervised visual representation learning,''
\newblock in {\em CVPR}, 2020, pp. 9729--9738.

\bibitem{Zhang_2021_CVPR}
Han Zhang, Jing~Yu Koh, Jason Baldridge, Honglak Lee, and Yinfei Yang,
\newblock ``Cross-modal contrastive learning for text-to-image generation,''
\newblock in {\em CVPR}, 2021, pp. 833--842.

\bibitem{10.1007/978-3-031-20047-2_25}
Zhixiong Pi, Weitao Wan, Chong Sun, Changxin Gao, Nong Sang, and Chen Li,
\newblock ``Hierarchical feature embedding for visual tracking,''
\newblock in {\em ECCV}, 2022, pp. 428--445.

\bibitem{Wu_2021_CVPR}
Qiangqiang Wu, Jia Wan, and Antoni~B. Chan,
\newblock ``Progressive unsupervised learning for visual object tracking,''
\newblock in {\em CVPR}, 2021, pp. 2993--3002.

\bibitem{Sandler_2018_CVPR}
Mark Sandler, Andrew Howard, Menglong Zhu, Andrey Zhmoginov, and Liang-Chieh Chen,
\newblock ``Mobilenetv2: Inverted residuals and linear bottlenecks,''
\newblock in {\em CVPR}, 2018.

\bibitem{Yan_2021_CVPR_Alpha_Refine}
Bin Yan, Xinyu Zhang, Dong Wang, Huchuan Lu, and Xiaoyun Yang,
\newblock ``Alpha-refine: Boosting tracking performance by precise bounding box estimation,''
\newblock in {\em CVPR}, 2021, pp. 5289--5298.

\bibitem{Fan_2019_CVPR}
Heng Fan, Liting Lin, Fan Yang, Peng Chu, Ge~Deng, Sijia Yu, Hexin Bai, Yong Xu, Chunyuan Liao, and Haibin Ling,
\newblock ``Lasot: A high-quality benchmark for large-scale single object tracking,''
\newblock in {\em CVPR}, 2019.

\bibitem{10.1007/978-3-319-10602-1_48}
Tsung-Yi Lin, Michael Maire, Serge Belongie, James Hays, Pietro Perona, Deva Ramanan, Piotr Doll{\'a}r, and C.~Lawrence Zitnick,
\newblock ``Microsoft coco: Common objects in context,''
\newblock in {\em ECCV}, 2014, pp. 740--755.

\bibitem{8922619}
Lianghua Huang, Xin Zhao, and Kaiqi Huang,
\newblock ``Got-10k: A large high-diversity benchmark for generic object tracking in the wild,''
\newblock {\em TPAMI}, vol. 43, no. 5, pp. 1562--1577, 2021.

\bibitem{Muller_2018_ECCV}
Matthias Muller, Adel Bibi, Silvio Giancola, Salman Alsubaihi, and Bernard Ghanem,
\newblock ``Trackingnet: A large-scale dataset and benchmark for object tracking in the wild,''
\newblock in {\em ECCV}, 2018.

\bibitem{Cao_2021_ICCV_HiFT}
Ziang Cao, Changhong Fu, Junjie Ye, Bowen Li, and Yiming Li,
\newblock ``Hift: Hierarchical feature transformer for aerial tracking,''
\newblock in {\em ICCV}, 2021, pp. 15457--15466.

\bibitem{Cao_2022_CVPR}
Ziang Cao, Ziyuan Huang, Liang Pan, Shiwei Zhang, Ziwei Liu, and Changhong Fu,
\newblock ``Tctrack: Temporal contexts for aerial tracking,''
\newblock in {\em CVPR}, 2022, pp. 14798--14808.

\bibitem{10225683}
{Cao, Ziang and Huang, Ziyuan and Pan, Liang and Zhang, Shiwei and Liu, Ziwei and Fu, Changhong},
\newblock ``Towards real-world visual tracking with temporal contexts,''
\newblock {\em TPAMI}, vol. 45, no. 12, pp. 15834--15849, 2023.

\bibitem{Kang_2023_ICCV}
Ben Kang, Xin Chen, Dong Wang, Houwen Peng, and Huchuan Lu,
\newblock ``Exploring lightweight hierarchical vision transformers for efficient visual tracking,''
\newblock in {\em ICCV}, 2023, pp. 9612--9621.

\bibitem{Blatter_2023_WACV}
Philippe Blatter, Menelaos Kanakis, Martin Danelljan, and Luc Van~Gool,
\newblock ``Efficient visual tracking with exemplar transformers,''
\newblock in {\em WACV}, 2023, pp. 1571--1581.

\bibitem{Danelljan_2019_CVPR_ATOM}
Martin Danelljan, Goutam Bhat, Fahad~Shahbaz Khan, and Michael Felsberg,
\newblock ``Atom: Accurate tracking by overlap maximization,''
\newblock in {\em CVPR}, 2019.

\bibitem{Yan_2021_ICCV}
Bin Yan, Houwen Peng, Jianlong Fu, Dong Wang, and Huchuan Lu,
\newblock ``Learning spatio-temporal transformer for visual tracking,''
\newblock in {\em ICCV}, 2021, pp. 10448--10457.

\bibitem{Cui_2022_CVPR}
Yutao Cui, Cheng Jiang, Limin Wang, and Gangshan Wu,
\newblock ``Mixformer: End-to-end tracking with iterative mixed attention,''
\newblock in {\em CVPR}, 2022, pp. 13608--13618.

\bibitem{Chen_2023_CVPR}
Xin Chen, Houwen Peng, Dong Wang, Huchuan Lu, and Han Hu,
\newblock ``Seqtrack: Sequence to sequence learning for visual object tracking,''
\newblock in {\em CVPR}, 2023, pp. 14572--14581.

\bibitem{7001050}
Yi~Wu, Jongwoo Lim, and Ming-Hsuan Yang,
\newblock ``Object tracking benchmark,''
\newblock {\em TPAMI}, vol. 37, no. 9, pp. 1834--1848, 2015.

\bibitem{10.1007/978-3-319-46448-0_27}
Matthias Mueller, Neil Smith, and Bernard Ghanem,
\newblock ``A benchmark and simulator for uav tracking,''
\newblock in {\em ECCV}, 2016, pp. 445--461.

\bibitem{sahin2023hoot}
Gozde Sahin and Laurent Itti,
\newblock ``Hoot: Heavy occlusions in object tracking benchmark,''
\newblock in {\em WACV}, 2023, pp. 4830--4839.

\end{thebibliography}

\end{document}